\documentclass[10pt, a4paper]{article}
\usepackage{lrec}
\usepackage{multibib}
\newcites{languageresource}{Language Resources}
\usepackage{graphicx}
\usepackage{tabularx}
\usepackage{soul}
\usepackage{cleveref}
\usepackage{amsmath}
\usepackage{multirow}
\usepackage{epstopdf}
\usepackage[latin1]{inputenc}

\usepackage{hyperref}
\usepackage{xstring}
\usepackage{makecell}
\usepackage{xspace}

\newcommand{\sys}{Train-o-Matic\xspace}

\title{Huge Automatically Extracted Training Sets for \\ Multilingual Word Sense Disambiguation}

\name{Tommaso Pasini, Francesco Elia, Roberto Navigli}

\address{Sapienza University of Rome\\
         \{pasini, elia, navigli\}@di.uniroma1.it}

\abstract{
We release to the community six large-scale sense-annotated datasets in multiple language to pave the way for supervised multilingual Word Sense Disambiguation. Our datasets cover all the nouns in the English WordNet and their translations in other languages for a total of millions of sense-tagged sentences . Experiments prove that these corpora can be effectively used as training sets for supervised WSD systems, surpassing the state of the art for low-resourced languages and providing competitive results for English, where manually annotated training sets are accessible. The data is available at \url{trainomatic.org}.
\newline \Keywords{Multilingual Word Sense Disambiguation, Resource, Dataset} }

\begin{document}

\maketitleabstract

\section{Introduction}
Word Sense Disambiguation is a crucial task in Natural Language Processing as it can be beneficial to several downstream applications, i.e., natural language understanding, semantic parsing and question answering. 
Despite the task has been around for a long time, it is far from being solved as it presents several challenges that have not fully been addressed yet, starting from the theoretical difficulty of formally establishing what a "word sense" is and choosing a corresponding sense inventory to the more pragmatic problems of finding large-scale sense-annotated corpora to train supervised systems on. Although WordNet \cite{Fellbaum:98}
virtually solved the first problem at least for English, a wide range of other issues still remain open. 
In fact, since supervised WSD systems need to be trained on a word-by-word basis, creating effective datasets requires a huge effort, which is beyond reach even for resource-rich languages like English. Clearly, this issue is even more severe for systems that need both lexicographic and encyclopedic knowledge \cite{schubert06} and/or need to work in a multilingual or domain-specific setting.
Knowledge-based WSD, on the other hand, exploits the knowledge contained in resources like WordNet to build algorithms (e.g. densest subgraph \cite{Moroetal:14tacl} or personalized page rank \cite{AgirreSoroa:09}) that can choose the sense of a word in context, thus not requiring training data but usually adopting bag-of-words approaches that neglect the lexical and syntactic context of the word (information that is more easily exploited by supervised systems), which may be essential in some scenarios. 
Furthermore, performances of both types of systems are highly affected by distribution of word senses that are usually different for each domain of application \cite{pasininavigli:18}.

In order to address these issues different solutions have been proposed in the past years, ranging from manually annotated resources that can be used to train WSD systems to automatic or semi-automatic approaches that aim at exploiting parallel corpora or partially annotated data in order 
to produce training corpora.
One of the first attempt to produce a sense annotated corpus is SemCor \citelanguageresource{Milleretal:93}, a collection of thousand sentences manually tagged with WordNet senses. 
While its quality is very high thanks to the effort of specialized annotators, it is far from covering the whole English vocabulary of words and senses. Moreover, such manual resources need extra effort to be maintained and updated to integrate new senses and words appearing in everyday language. Thus, in order to overcome these issues, semi-automatic or fully automatic approaches have been proposed over the past years.

\citelanguageresource{taghipourng:15} exploit a parallel corpus and the manual translations of senses to annotate the words in the corpus with senses. Similarly, but without the need for human intervention, \newcite{dellibovietal:17} and \citelanguageresource{camachocolladosetal:16}, rely on aligned sentences in order to create a richer context that can be beneficial to their disambiguation. 
\newcite{raganatoetal:16}, instead, designed a set of heuristics which exploit the human effort of the Wikipedia community in order to propagate and add sense annotations to the Wikipedia pages.
Similarly \newcite{pasininavigli:17} exploit a knowledge base in order to annotate sentences with sense tags and uses a measure of confidence in order to select the most correct annotated sentences. 
They show that, relying on a multilingual semantic network as the underlying knowledge base, they are able to create high-quality sense-tagged corpora for any languages supported by the semantic network.

Our work builds upon that of \newcite{pasininavigli:17} in order to generate sense-tagged corpora for 5 major European languages (English, French, German, Spanish and Italian) and the most spoken language of Asia (Chinese) and paves the way for supervised Word Sense Disambiguation in multiple languages.
Exploiting the knowledge contained in BabelNet \cite{NavigliPonzetto:10,NavigliPonzetto:12} -- a huge and multilingual semantic network containing both lexicographic and encyclopedic knowledge -- and Wikipedia, we generated large corpora annotated with BabelNet senses for the 6 languages listed above.\\
Experiments and statistics prove that these automatically created corpora are rich in terms of number of different lemmas annotated with a sense and number of sentences, and as such they can be a valuable resource for supervised WSD systems: in fact, systems trained on our datasets perform better or comparably to the state of the art across different languages. The added value is even more visible on low-resourced languages where such data is very scarce, if at all available.
We now give an overview of our corpus building procedure, including a brief description of \sys; we then discuss features of the created datasets, our experimental setup for evaluation and its results.

\section{Building the corpus}
In order to build a sense annotated corpus for a given language $L$, our system takes as input a corpus of raw sentences $C$ in the language $L$, a list of words $W_L$ in the target language $L$ and a semantic network $G$\footnote{We consider a WordNet-like structure of the semantic network, where the nodes are synsets (concepts) which contain a set of lemmas that can express that concept.}.
For each language $L$ we apply \cite{pasininavigli:17}[\sys] in order to annotate each target word $w \in W_L$ with a distribution over its senses.

For example given the ambiguous sentence \textit{"A match is a lighter."} and the target word \textit{"match"}, \sys will output a sense distribution of the target word similar to the following:\\
$$[\text{match}_n^1:0.74, \text{match}_n^2:0.16, \text{match}_n^3:0.10]$$
where $word_{pos}^{n}$ follows the notation introduced in \cite{navigli:09} to indicate the \textit{n-th} WordNet sense of $word$ with Part-of-Speech $pos$. 

We chose Wikipedia in the language $L$ as raw corpus $C_L$ and BabelNet as the underlying semantic graph $G$ because both are available for all the 6 languages of interest. BabelNet is also exploited in order to generate the lexicon $W_L$ for each language $L$ by collecting all the lexicaliztions of a synset in the graph in the given language $L$.
Given the size of BabelNet we chose not to include all of its synsets, limiting our graph only to those that contain at least a sense from WordNet.
We choose to keep all the BabelNet edges because they add many syntagmatic relations on top of the manually curated paradigmatic edges of WordNet.

To build each corpus we select all the sentences in each Wikipedia that contain at least one of the target words in $W_L$ and then apply \sys.

\subsection{\sys Overview}
\label{sec:systemoverview}
\sys is a 3-step method to annotate a raw corpus of sentences.

\paragraph{1. Lexical Profiling}
\sys exploits the semantic graph $G$ in order to generate a lexical profile for each of the synsets in $G$.
Such profile is computed by running the Personalized PageRank algorithm \cite{BrinPage:98} for each node in the graph.
This means that, given the following formula:
\begin{equation}
\label{eqn:ppr}
v^{(t+1)} = (1-\alpha) v^{(0)} + \alpha M v^{(t)}
\end{equation}

we set a $1$ in the probability distribution $v$ to the component that corresponds to the node for which we want to build the lexical profile.
This procedure can also be interpreted as a random walk on the graph $G$ where the walk is always restarted from the same initial node.

At the end of this step each synset $s$ (i.e. node) in the graph has an associated vector in which each component represents another synset $s'$ in the graph and the value of the component expresses the probability of reaching $s'$ from $s$; this probability can be interpreted as a measure of relatedness between $s$ and $s'$.

\paragraph{2. Sentence Scoring}
Once we have a distribution over the most related concepts for each synset in the graph, \sys exploits them in order to annotate each target word in the raw corpus.
For example, given the target word $w = \text{"match"}$, its set of senses retrieved from the semantic network $S_\text{match} =
[match_n^1, match_n^2]$  and the sentence \textit{"Messi didn't play the last match."} which contains the target word, the system creates a distribution over the senses in $S_\text{match}$.

This is done by approximating the probability of a sense given the target word and the sentence as follows:

\begin{align}
     \label{sentencebayes}
          P(s | \sigma,w) &= \frac{P(\sigma | s,w) P(s|w)}{P(\sigma|w)} \\
        &\label{eqn:indep} \approx P(w_1|s,w) \dots P(w_n|s,w) P(s|w)
    \end{align}

which assumes the independence of the words and removes the constant denominator. 
Each probability in (\ref{eqn:indep}) is computed exploiting the vectors previously computed.
In fact, grounding the formula on our example, we have:
\begin{align}
    &P(\text{match}_n^1 | \text{Messi didn't play the last match}, \text{match}) = \\
    & P(\text{match}|\text{match}_n^1, \text{match}) \times\\
    & P(\text{play} | \text{match}_n^1, \text{match}) \times \\
    & P(\text{Messi} | \text{match}_n^1, \text{match})
\end{align}
and each individual probability for the words $w_i$ is computed by taking the value of the synset with the highest probability in the lexical profile of $match_n^1$ that contains the lemma $w_i$.
\paragraph{3. Sentence Ranking}
The last step aims at sorting and removing the sentences which are less likely to be correctly tagged. The sentences are in fact ranked by a confidence score which is computed by considering the difference between the most likely and second most likely senses of the target word. For example, referring to the previous example sentence, if $\text{match}_n^1$ received a probability of $.7$ and $\text{match}_n^2$ one of $.3$ then the sentence score will be $.4$.
For each sense of a given word $w$, the candidate sentences are sorted using the confidence score. In order to select how many sentences to include in total, we set a parameter $K$ that represents how many sentences must be included for the first sense of the given target word (i.e., the most common sense), with subsequent senses (according to the BabelNet ordering) for the same word receiving a decreasing number of examples computed according to a Zipf's distribution. 
\begin{table*}[ht]
    \centering
    \resizebox{\textwidth}{!}{
    \begin{tabular}{|l|r||r|r|r|r|r|r|}
    \cline{2-8}
    \multicolumn{1}{l|}{} & \multicolumn{1}{c||}{Total} & \multicolumn{1}{c|}{English}& \multicolumn{1}{c|}{French} &
    \multicolumn{1}{c|}{German} & \multicolumn{1}{c|}{Italian} & \multicolumn{1}{c|}{Spanish} & \multicolumn{1}{c|}{Chinese} \\
    \hline
    \multicolumn{1}{|c|}{\makecell{Number of \\Annotations}} & 17,987,488 & 12,722,530 & 1,597,230 & 1,213,634 & 1,037,253 & 935,713 & 481,128\\
    \hline
     \makecell{Distinct lemmas \\covered} & 146,068 & 51,395 & 25,689 & 22,300 & 19,192 & 14,596 & 12,896 \\
    \hline
     \makecell{Distinct senses\\ covered} & 63,613 & 56,229 & 33,843 & 23,526 & 22,587 & 21,388 & 12,485\\
    \hline
     \makecell{Average \# of sentences \\per sense} & 75.5 & 226.3 & 47.2 & 51.6 & 45.9 & 43.7 & 38.5 \\
    \hline
     \makecell{Average confidence \\score} & 56.74 & 71.64 & 22.07 & 89.19 & 19.40 & 50.41 & 87.75\\
    \hline
    \makecell{Average Polisemy}& 3.57 & 1.57 & 7.29 & 1.13 & 4.98 & 5.07 & 1.41\\
    \hline
    \end{tabular}
    }
    \caption{Statistics for each corpus in each language.}
    \label{tab:corporastats}

\end{table*}

\begin{table}[ht]
    \centering
    \resizebox{\linewidth}{!}{
    \begin{tabular}{|r|r|r|r|}
    \hline
        Corpus & Sentences & Annotations & Unique Words \\
        \hline
        \hline
        SemCor & 37,176 & 226,036 & 22,436\\
        SemCor+OMSTI & 850,974 & 1,137,170 & 22,437\\
        \sys & 12,722,530 & 12,722,530 & 51,395 \\
        \hline
    \end{tabular}
    }
    \caption{Statistics of SemCor, OMSTI and \sys about the number of sentences, annotations and unique words.}
    \label{tab:corporacomp}
    
\end{table}
The following formula better explains the computation of the number of sentences assigned to each sense in a given ordering $o$.

$$
\text{examples}_s = K \times \frac{1}{index(o, s)^z}
$$

where $index(o, s)$ is a function that returns the position of a synset $s$ in the ordering $o$.
So, for example, if $K$ is set to $100$, the first sense of the target word will receive $K$ examples, the second one $\frac{K}{2^z}$ and so on; $z$ is another parameter of the system. 


\section{Statistics}
In this section we report some features of the corpora produced by \sys, in order to give a complete overview of the data.

In Table \ref{tab:corporastats} we show the number of annotations for each language as well as the number of distinct words and senses that have at least one example in our corpora and the number of sentences for each sense on average. 

\sys was able to generate around $18M$ annotated sentences for roughly $146K$ distinct lemmas and $63K$ distinct senses across languages. These corpora proved also to be of high quality, taking supervised system on par with or beyond state of the art results (Section \ref{sec:results}). 
The number of annotations is bigger for English and comparable across other languages: this is both because, for English, we set the value of the parameter $K$ (see Section \ref{sec:systemoverview}) to $500$ instead of $100$, and because BabelNet, on average, contains more English senses compared to other languages.

As can be seen, each language has an average of $75$ different sentences for each sense in the corpus, with English having the highest number of sentences per sense. 
Note that the total number of distinct senses covered is not equal to the sum of distinct senses for each sense due to the fact that we use a language-independent sense inventory (i.e. BabelNet) similarly to \newcite{otegietal:16} and \newcite{dellibovietal:17}. Thus many senses are shared across languages.
The average confidence score measures how confident the system was on average when annotating the given language, meaning that the resulting data is most likely better: this score depends on both the average ambiguity of each lemma and on the quality of the relations in BabelNet. 
As expected, the system confidence score is highest in languages that have the lowest polisemy, i.e. English and German, which have the lowest average number of senses for nouns.
As regards the average number of sentences for each sense, it directly depends on the parameter $K$ and $z$ that we set experimentally (see Section \ref{sec:systemoverview}). All corpora but English proved to lead supervised system to better performance when $K$ was set to $100$ and z between $2.0$ and $3.0$, thus we preferred to keep a lower number of more accurate sentences (50 for each sense).
The English corpus, instead, was generated with $K$ equal to $500$ and z equal to $2.0$ and thus it has a higher average number of sentences for each sense.

Table \ref{tab:corporacomp}, instead, shows the comparison, in terms of number of sentences, annotations and unique words covered, between our automatically generated English corpus and two other corpora:
\begin{itemize}
    \item SemCor (Miller et al., 1993), a corpus containing about 226,000 tokens annotated manually with WordNet senses.
    \item One Million Sense-Tagged Instances (Taghipour and Ng, 2015)[SemCor+OMSTI], a sense-annotated dataset obtained via a semi-automatic approach based on the disambiguation of a parallel corpus, i.e., the United Nations Parallel Corpus, performed by exploiting manually translated word senses. It also contains SemCor.
\end{itemize}
In terms of number of annotated sentences and number of annotations, our corpus is significantly bigger than SemCor and SemCor+OMSTI (by a factor of 200 and 10 respectively). More importantly, however, it covers double the amount of nouns that are covered by these two corpora, allowing supervised systems to have higher recall and to rely less on the Most Frequent Sense heuristic.

\begin{table*}[ht!]
\centering
\begin{tabular}{|l |l | r | r | r | r|} 
\hline
 \multirow{2}{*}{Test Set} & \multirow{2}{*}{Language} & \multicolumn{3}{c|}{\sys} & \multicolumn{1}{c|}{Best System}\\
 \cline{3-6}
 & & Precision & Recall & F1 & F1\\
\hline
\hline
\multirow{4}{*}{SemEval 2013} & German & 0.66 & 0.61 & \textbf{0.63} & 0.62\\
& French & 0.61 & 0.60 & \textbf{0.61} & \textbf{0.61}\\
& Spanish & 0.68 & 0.66  & 0.67 & \textbf{0.71}\\
& Italian & 0.71     & 0.66  &\textbf{ 0.68} & 0.66 \\
\hline
\hline
\multirow{2}{*}{SemEval 2015} & Spanish & 61.3 & 54.8 & \textbf{57.9} &  56.3 \\ 
& Italian & 65.1 & 55.6 & \textbf{59.9} & 56.6  \\ 

\hline
\end{tabular}
\caption{Precision, Recall and F1 of IMS trained on \sys, against the best performing system on SemEval-13 and SemEval-15.}

\label{tab:multilingualresults}
\end{table*}
\begin{table}[ht!]
\centering
\resizebox{\linewidth}{!}{
\begin{tabular}{|l | r | r | r | r |} 
\hline
Dataset & \sys & OMSTI & SemCor & MFS \\
\hline
\hline
Senseval-2 & 70.5 & 74.1 & \textbf{76.8} & 72.1\\
Senseval-3 &67.4 & 67.2 &  \textbf{73.8} & 72.0\\
SemEval-07 & 59.8 & 62.3 & \textbf{67.3} & 65.4\\
SemEval-13 & \textbf{65.5} & 62.8 & \textbf{65.5} & 63.0\\
SemEval-15 & \textbf{68.6} & 63.1 & 66.1 & 66.3\\
\hline
\hline
ALL & 67.3 & 66.4 & \textbf{70.4} & 67.6 \\
\hline

\end{tabular}
}
\caption{F1 of IMS trained on \sys, OMSTI and SemCor, and MFS for the Senseval-2, Senseval-3, SemEval-07, SemEval-13 and SemEval-15 datasets.}

\label{table:mfsalldataset}
\end{table}
\section{Experimental Setup}
In order to evaluate the quality of the corpora we tested the performance of IMS, a state-of-the-art WSD system, when trained on our datasets.

\paragraph{English setup:} For English, we compare the performance of IMS when trained on \sys to that obtained against training with: OMSTI and SemCor.

The evaluation has been performed using the unified evaluation framework for Word Sense Disambiguation made available by \newcite{raganatoetal:17}, thus considering the following WSD shared tasks: Senseval-2 \cite{EdmondsCotton:01}, Senseval-3 \cite{SnyderPalmer:04}, SemEval-2007 \cite{naviglietal:07}, SemEval-2013 \cite{Naviglietal:13semeval} and SemEval-2015 \cite{MoroNavigli:15}.
We set the two \sys parameters $K$ to $500$ and $z$ to $2.0$ experimentally, testing the models learned by IMS on a small in-house development set\footnote{The development set contains roughly 50 items per language.} and choosing the one with the highest performance.

\paragraph{Multilingual setup:}
For the other languages we tuned the two paramenter $K$ and $z$ in the same way we did for English. The corpora proved to be more effective with $K$ set to 100, for all the languages, and $z$ ranging in $[2.0,3.0]$.
To prove that the generated data in the other languages are also high quality we also report the performance of IMS when trained on \sys corpora for Italian and Spanish on the Multilingual WSD task of SemEval-2015 \cite{MoroNavigli:15}, and for German, French, Spanish and Italian on the Multilingual WSD task of SemEval-2013 \cite{Naviglietal:13semeval} which focuses on nouns only.
Given that no supervised system have been submitted to this task\footnote{Note that no supervised system have ever been submitted for a multilingual WSD task.} we compare against the best performing knowledge-based systems of the two SemEvals.

\subsection{Results}
\label{sec:results}

 \paragraph{English results:} As can be seen in Table \ref{table:mfsalldataset} IMS trained on our corpus is always comparable, if not better (from 2 to 3 points), than OMSTI\footnote{We recall that OMSTI has been built using a semi-automatic approach and contains SemCor}. SemCor, instead, provides better training data for 3 out of 5 datasets, while the performance of IMS is comparable on the SemEval-2013 and SemEval-2015.
 %
 This shows that our automatically generated data can lead to better performance than semi-automatic datasets and, in some situations, even surpass that of manually annotated ones.
 More interestingly, the ability to automatically generate high-quality sense-annotated data enables the creation of domain-specific datasets that could be used to train WSD systems on particular domains of interest. Given that such a system would most likely outperform a system trained on non-specialized data (e.g. because the latter may have learned a Most Frequent Sense bias that is not accurate for the domain at hand), this is often a need for companies which need to specialize their software on a specific use case (see \cite{pasininavigli:17} for experiments on domain specific tasks).
 
 \paragraph{Multilingual results:}
 Looking now at results in Table \ref{tab:multilingualresults} it is clear that the best improvement in performance, compared to the current state of the art, is obtained on low-resourced languages, which was our main objective.
 We note that IMS, when trained on \sys corpora, is able to score from $1$ to $3$ points more than the best system of each language and each of the two SemEval (i.e. SemEval 2013 and SemEval 2015) but Spanish in SemEval 2013. 
 
 This comes as expected as supervised systems perform better than knowledge-based ones \cite{raganatoetal:17} when enough training data is available. 
 Still, it is not the purpose of this paper to show that these datasets provide the best possible training sets in all scenarios, but rather that they can be very valuable in low-resourced languages, for which training supervised systems would be otherwise impossible.

\section{Conclusion}
We release to the community 6 sense-annotated corpora for the 5 major European languages (English, French, Spanish, German and Italian) plus Chinese, each containing on average more than 1 million sentences from Wikipedia articles and automatically annotated using \sys. 

Our experiments proved that these corpora provide effective training ground for supervised WSD system, especially in a multilingual setting where sense annotated data is scarce, if at all available. As a matter of fact, the performance of supervised systems trained on this data is better or comparable to those trained on semi-automatically and, in some cases, manually-curated data.
Given the lack of such data for languages other than English, most WSD systems that target these languages usually adopt a knowledge-based approach, thus neglecting syntactic and contextual information that may be essential in some scenarios. This point is confirmed by the fact that we are able to outperform such systems by using these corpora as training set. All these points show that our corpora are able to address the need for sense-annotate data in low-resources languages. 
The data is available at \url{trainomatic.org}.

\section*{Acknowledgments}
\noindent
\begin{minipage}{0.1\linewidth}
    \raisebox{-0.2\height}{\includegraphics[trim =32mm 55mm 30mm 5mm, clip, scale=0.2]{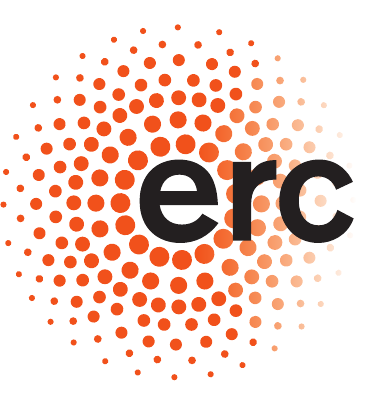}}
\end{minipage}
\hspace{0.01\linewidth}
\begin{minipage}{0.72\linewidth}
  The authors gratefully acknowledge the support of the ERC Consolidator Grant MOUSSE No. 726487.
\end{minipage}
\hspace{0.01\linewidth}
\begin{minipage}{0.05\linewidth}
\raisebox{-0.25\height}{\includegraphics[trim =0mm 5mm 5mm 2mm,clip,scale=0.078]{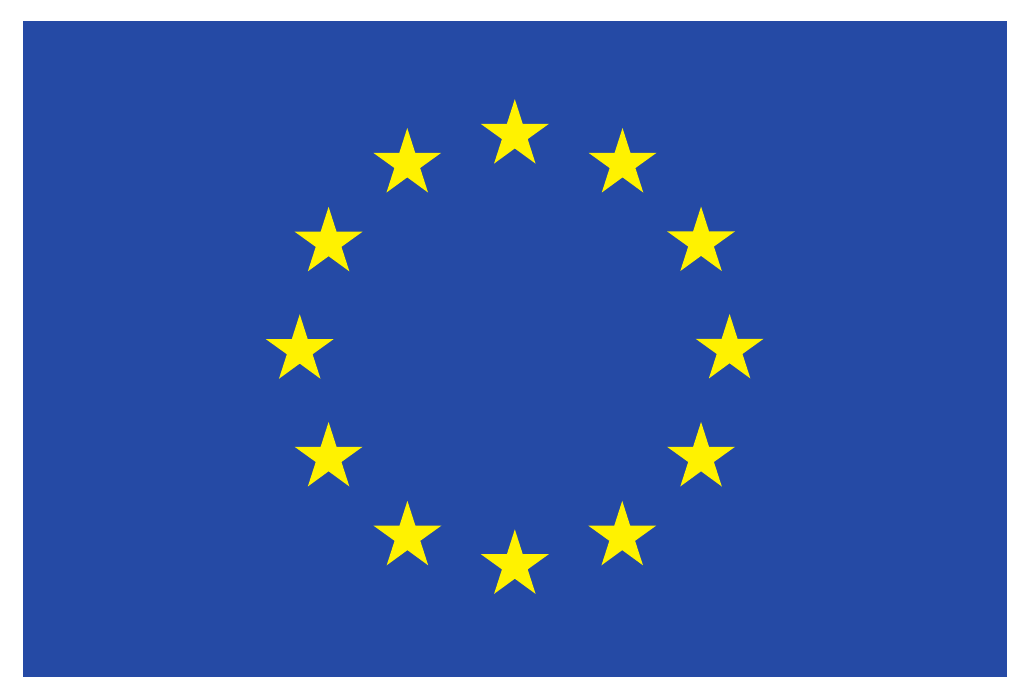}}
\end{minipage}
\section{Bibliographical References}
\label{main:ref}

\bibliographystyle{lrec}
\bibliography{bibliography}

\begin{thebibliography}{}

\bibitem[\protect\citename{Camacho-Collados \bgroup et al.\egroup
  }2016]{camachocolladosetal:16}
Jos\'{e} Camacho-Collados and Claudio Delli Bovi and Alessandro Raganato and
  Roberto Navigli.
\newblock (2016).
\newblock {\em {A Large-Scale Multilingual Disambiguation of Glosses}}.

\bibitem[\protect\citename{Miller \bgroup et al.\egroup }1993]{Milleretal:93}
Miller, George A. and Leacock, Claudia and Tengi, Randee and Bunker, Ross.
\newblock (1993).
\newblock {\em A Semantic Concordance}.

\bibitem[\protect\citename{Taghipour and Ng}2015]{taghipourng:15}
Taghipour, Kaveh and Ng, Hwee Tou.
\newblock (2015).
\newblock {\em One Million Sense-Tagged Instances for Word Sense Disambiguation
  and Induction}.
\newblock Association for Computational Linguistics.

\end{thebibliography}


\begin{thebibliography}{}

\bibitem[\protect\citename{Agirre and Soroa}2009]{AgirreSoroa:09}
Agirre, E. and Soroa, A.
\newblock (2009).
\newblock Personalizing {P}age{R}ank for {W}ord {S}ense {D}isambiguation.
\newblock In {\em Proceedings of the 12th Conference of the European Chapter of
  the Association for Computational Linguistics, {Athens, Greece, 30 March--3
  April 2009}}, pages 33--41.

\bibitem[\protect\citename{Brin and Page}1998]{BrinPage:98}
Brin, S. and Page, L.
\newblock (1998).
\newblock The anatomy of a large-scale hypertextual web search engine.
\newblock {\em Computer Networks and ISDN Systems}, 30(1--7):107--117.

\bibitem[\protect\citename{Delli~Bovi \bgroup et al.\egroup
  }2017]{dellibovietal:17}
Delli~Bovi, C., Camacho-Collados, J., Raganato, A., and Navigli, R.
\newblock (2017).
\newblock {EuroSense}: Automatic harvesting of multilingual sense annotations
  from parallel text.
\newblock In {\em Proc.of ACL}, volume~2, pages 594--600.

\bibitem[\protect\citename{Edmonds and Cotton}2001]{EdmondsCotton:01}
Edmonds, P. and Cotton, S.
\newblock (2001).
\newblock Senseval-2: overview.
\newblock In {\em Proc. of SensEval 2}, pages 1--5. ACL.

\bibitem[\protect\citename{Fellbaum}1998]{Fellbaum:98}
Christiane Fellbaum, editor.
\newblock (1998).
\newblock {\em {W}ord{N}et: An Electronic Database}.
\newblock MIT Press, Cambridge, MA.

\bibitem[\protect\citename{Moro and Navigli}2015]{MoroNavigli:15}
Moro, A. and Navigli, R.
\newblock (2015).
\newblock Semeval-2015 task 13: Multilingual all-words sense disambiguation and
  entity linking.
\newblock In {\em Proc. of SemEval-2015}.

\bibitem[\protect\citename{Moro \bgroup et al.\egroup }2014]{Moroetal:14tacl}
Moro, A., Raganato, A., and Navigli, R.
\newblock (2014).
\newblock {Entity Linking meets Word Sense Disambiguation: a Unified Approach}.
\newblock {\em Transaction of ACL (TACL)}, 2:231--244.

\bibitem[\protect\citename{Navigli and Ponzetto}2010]{NavigliPonzetto:10}
Navigli, R. and Ponzetto, S.~P.
\newblock (2010).
\newblock {B}abel{N}et: Building a very large multilingual semantic network.
\newblock In {\em Proceedings of the 48th Annual Meeting of the Association for
  Computational Linguistics, {Uppsala, Sweden, 11--16 July 2010}}, pages
  216--225.

\bibitem[\protect\citename{Navigli and Ponzetto}2012]{NavigliPonzetto:12}
Navigli, R. and Ponzetto, S.~P.
\newblock (2012).
\newblock {B}abel{N}et: {T}he automatic construction, evaluation and
  application of a wide-coverage multilingual semantic network.
\newblock {\em Artificial Intelligence}, 193:217--250.

\bibitem[\protect\citename{Navigli \bgroup et al.\egroup }2007]{naviglietal:07}
Navigli, R., Litkowski, K.~C., and Hargraves, O.
\newblock (2007).
\newblock Semeval-2007 task 07: Coarse-grained {E}nglish all-words task.
\newblock In {\em Proceedings of the 4th International Workshop on Semantic
  Evaluations (SemEval-2007), {Prague, Czech Republic, 23--24 June 2007}},
  pages 30--35.

\bibitem[\protect\citename{Navigli \bgroup et al.\egroup
  }2013]{Naviglietal:13semeval}
Navigli, R., Jurgens, D., and Vannella, D.
\newblock (2013).
\newblock Semeval-2013 task 12: Multilingual word sense disambiguation.
\newblock In {\em Second Joint Conference on Lexical and Computational
  Semantics (* SEM '13)}, volume~2, pages 222--231.

\bibitem[\protect\citename{Navigli}2009]{navigli:09}
Navigli, R.
\newblock (2009).
\newblock {W}ord {S}ense {D}isambiguation: {A} survey.
\newblock {\em ACM Computing Surveys}, 41(2):1--69.

\bibitem[\protect\citename{Otegi \bgroup et al.\egroup }2016]{otegietal:16}
Otegi, A., Aranberri, N., Branco, A., Hajic, J., Neale, S., Osenova, P.,
  Pereira, R., Popel, M., Silva, J., Simov, K., et~al.
\newblock (2016).
\newblock Qtleap wsd/ned corpora: Semantic annotation of parallel corpora in
  six languages.
\newblock In {\em Proceedings of the 10th Language Resources and Evaluation
  Conference, LREC}, pages 3023--3030.

\bibitem[\protect\citename{Pasini and Navigli}2017]{pasininavigli:17}
Pasini, T. and Navigli, R.
\newblock (2017).
\newblock Train-o-matic: Large-scale supervised word sense disambiguation in
  multiple languages without manual training data.
\newblock In {\em Proceedings of the 2017 Conference on Empirical Methods in
  Natural Language Processing}, pages 78--88. Association for Computational
  Linguistics.

\bibitem[\protect\citename{Pasini and Navigli}2018]{pasininavigli:18}
Pasini, T. and Navigli, R.
\newblock (2018).
\newblock Two knowledge-based methods for high-performance sense distribution
  learning.
\newblock In {\em AAAI-2018}.

\bibitem[\protect\citename{Raganato \bgroup et al.\egroup
  }2016]{raganatoetal:16}
Raganato, A., {Delli Bovi}, C., and Navigli, R.
\newblock (2016).
\newblock {Automatic Construction and Evaluation of a Large Semantically
  Enriched Wikipedia}.
\newblock In {\em Proceedings of IJCAI}, pages 2894--2900, New York City, NY,
  USA, July.

\bibitem[\protect\citename{Raganato \bgroup et al.\egroup
  }2017]{raganatoetal:17}
Raganato, A., Camacho-Collados, J., and Navigli, R.
\newblock (2017).
\newblock {Word Sense Disambiguation: A Unified Evaluation Framework and
  Empirical Comparison}.
\newblock In {\em Proc. of EACL}, pages 99--110, Valencia, Spain.

\bibitem[\protect\citename{Schubert}2006]{schubert06}
Schubert, L.~K.
\newblock (2006).
\newblock Turing's dream and the knowledge challenge.
\newblock In {\em Proc. of AAAI-06}, pages 1534--1538.

\bibitem[\protect\citename{Snyder and Palmer}2004]{SnyderPalmer:04}
Snyder, B. and Palmer, M.
\newblock (2004).
\newblock The english all-words task.
\newblock In {\em Proc. of Senseval-3}, pages 41--43, Barcelona, Spain.

\end{thebibliography}

\section{Language Resource References}
\label{lr:ref}
\bibliographystylelanguageresource{lrec}
\bibliographylanguageresource{language_resources}

\end{document}